\pdfoutput=1

\documentclass[11pt]{article}

\usepackage{ACL2023}

\usepackage{times}
\usepackage{latexsym}

\usepackage[T1]{fontenc}

\usepackage[utf8]{inputenc}

\usepackage{microtype}

\usepackage{inconsolata}

\usepackage{xspace}
\usepackage{graphicx}
\usepackage{amssymb}
\usepackage{multirow}
\usepackage{multicol}
\usepackage{array}
\usepackage{booktabs}
\usepackage{algorithm}
\usepackage{algorithmic}
\usepackage{amsmath}
\usepackage{makecell}
\usepackage{amsfonts}
\usepackage{bbold}
\usepackage{amsthm,amsmath,amssymb} 

\DeclareMathAlphabet\mathbfcal{OMS}{cmsy}{b}{n}

\title{Commonsense Knowledge Graph Completion \\Via Contrastive Pretraining and Node Clustering}

\author{Siwei Wu, Xiangqing Shen, and Rui Xia\thanks{*Corresponding author} \\
        School of Computer Science and Engineering, \\ Nanjing University of Science and Technology, China \\
        \texttt{\{wusiwei, xiangqing.shen, rxia\}@njust.edu.cn}}

\begin{document}
\maketitle
\begin{abstract}

The nodes in the commonsense knowledge graph (CSKG) are normally represented by free-form short text (e.g., word or phrase).
Different nodes may represent the same concept. 
This leads to the problems of edge sparsity and node redundancy, which challenges CSKG representation and completion. On the one hand, edge sparsity limits the performance of graph representation learning; On the other hand, node redundancy makes different nodes corresponding to the same concept have inconsistent relations with other nodes. To address the two problems, we propose a new CSKG completion framework based on Contrastive Pretraining and Node Clustering (CPNC). 
Contrastive Pretraining constructs positive and negative head-tail node pairs on CSKG and utilizes contrastive learning to obtain better semantic node representation. 
Node Clustering aggregates nodes with the same concept into a latent concept, assisting the task of CSKG completion.
We evaluate our CPNC approach on two CSKG completion benchmarks (CN-100K and ATOMIC), where CPNC outperforms the state-of-the-art methods.
Extensive experiments demonstrate that both Contrastive Pretraining and Node Clustering can significantly improve the performance of CSKG completion.
The source code of CPNC is publicly available on \url{https://github.com/NUSTM/CPNC}.

\end{abstract}

\section{Introduction}

\begin{figure}[!htp]
    \centering
    \includegraphics[width=0.98\columnwidth]{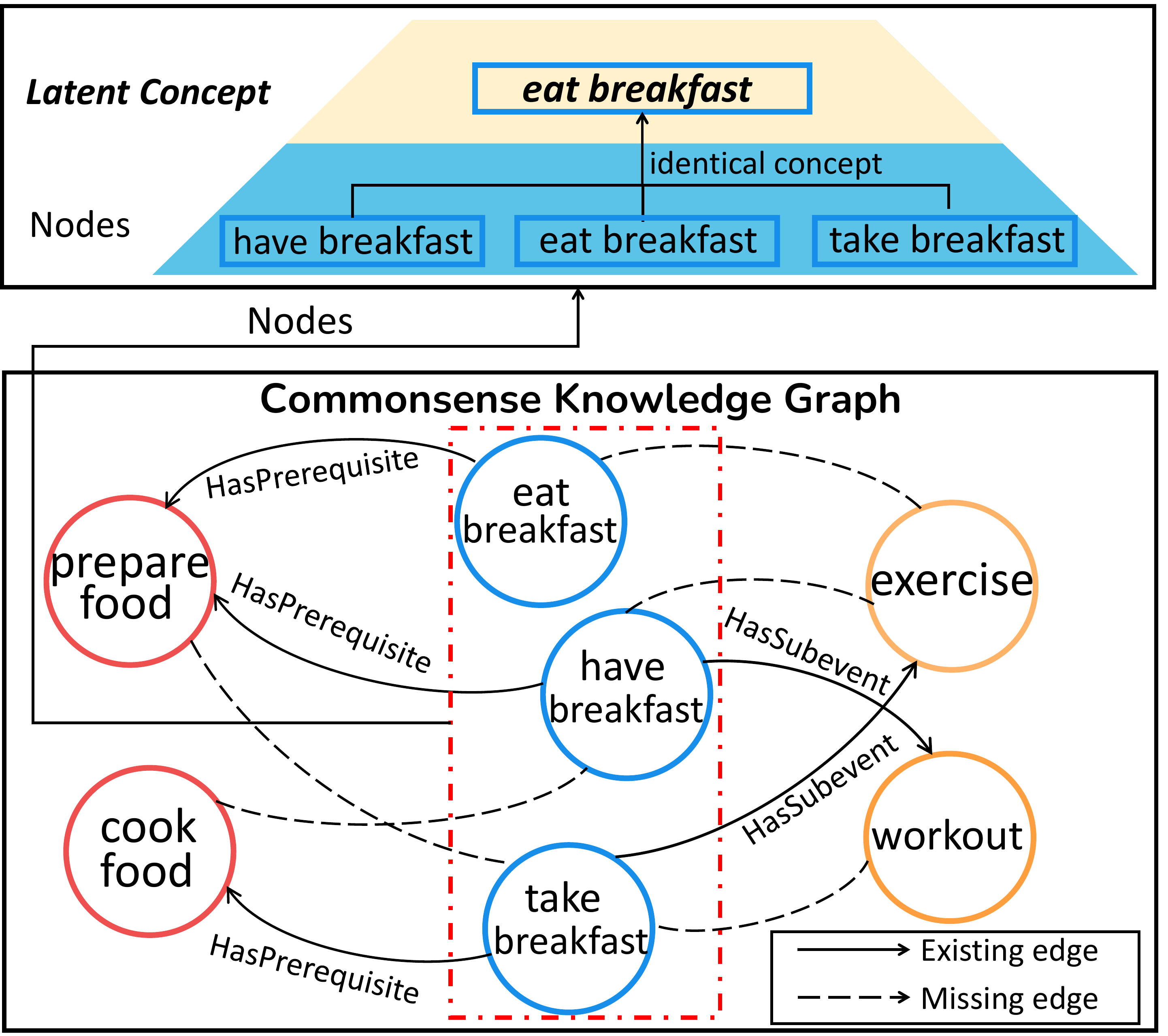}
    \caption{Illustration of a subgraph and the latent concept in ConceptNet. The nodes of the same color belong to the same latent concept.}
    \label{fig:explaination}
\end{figure}

Commonsense knowledge graphs (CSKG) have been widely used to build commonsense-grounded AI applications, such as question answering~\citep{DBLP:conf/aaai/LvGXTDGSJCH20}, visual question answering~\citep{DBLP:conf/ijcai/ZhuYWS0W20}, sentiment analysis~\citep{DBLP:conf/ijcai/LiM0L0C0Z22}, dialogue system~\cite{DBLP:conf/acl/TuLC0W022}, etc.
Commonsense knowledge graphs such as ConceptNet~\cite{DBLP:conf/aaai/SpeerCH17} and ATOMIC~\cite{DBLP:conf/aaai/SapBABLRRSC19} provide a structured way of representing a commonsense concept, which consists of a head node, a tail node, and the relation edge.
Nodes in commonsense knowledge graphs are typically represented by free-form short text (word or phrase), resulting in many different nodes representing the same concept.
Figure~\ref{fig:explaination} shows a subgraph of ConceptNet, where nodes in the same color indicate the same concept.
E.g., ``\emph{have breakfast}'', ``\emph{take breakfast}'' and ``\emph{eat breakfast}'' all express the concept of ``\textbf{\emph{eat breakfast}}''.
This problem also results in a large number of missing edges between nodes, as illustrated by the dashed lines in Figure~\ref{fig:explaination}.
E.g., ``\emph{eat breakfast}'' and ``\emph{prepare food}'' have a ``\emph{HasPrerequisite}'' relation, but such relation is missing between ``\emph{eat breakfast}'' and ``\emph{cook food}''.

On the one hand, node redundancy in CSKG makes different nodes of the same concept have inconsistent relations with other nodes. 
In CSKG representation and completion, it would be beneficial to make use of the latent concept information behind different nodes to help learn more semantic-general representations.
However, this intuition was ignored by most of the existing work in CSKG completion.
On the other hand, as analyzed by~\citet{DBLP:conf/aaai/MalaviyaBBC20}, edge sparsity in CSKG limits information propagation in graph neural networks.
Upon graph neural networks, researchers further incorporated pre-trained language model such as BERT to enhance the semantic representation of nodes~\cite{DBLP:conf/aaai/MalaviyaBBC20,DBLP:conf/cikm/JuYL22,DBLP:conf/ijcnn/WangWHYLK21}.
However, fine-tuning BERT is still imperfect in representing
the commonsense knowledge graph, which is made up of linked nodes represented in a free-form short text.

To tackle the two issues, we propose a new CSKG completion framework based on the Encoder-Decoder architecture, which contains two core modules \emph{\textbf{C}ontrastive \textbf{P}retraining and \textbf{N}ode \textbf{C}lustering} (CPNC).
Contrastive Pretraining is to alleviate the difficulty of node representation learning induced by edge sparsity. 
Through contrastive learning on positive and negative head-tail node pairs, the embedding distance between related nodes becomes closer, and that between unrelated nodes becomes farther so as to learn better node representations.
Node Clustering aims to address the edge inconsistency issue caused by node redundancy.
We cluster nodes with close semantic representations and take the mean vector as the latent concept representation for nodes in this cluster.
To assist CSKG completion, the latent concept representation is fused with the node representation.

We evaluate our CPNC framework on two CSKG completion benchmarks, i.e., CN-100K and ATOMIC. 
The results show that our model outperforms the state-of-the-art models for this task significantly. 
Ablation studies demonstrate that both Contrastive Pretraining and Node Clustering modules can significantly improve the performance of CSKG completion.
Further experiments verify that our model can consistently improve as the sparsity of knowledge graphs increases: the higher the sparsity, the more improvement our model achieves.

\section{Related Work}
We briefly review traditional knowledge graph completion, and then pay more attention to commonsense knowledge graph completion.

\begin{figure*}[!htp]
    \centering
    \includegraphics[width=0.99\textwidth]{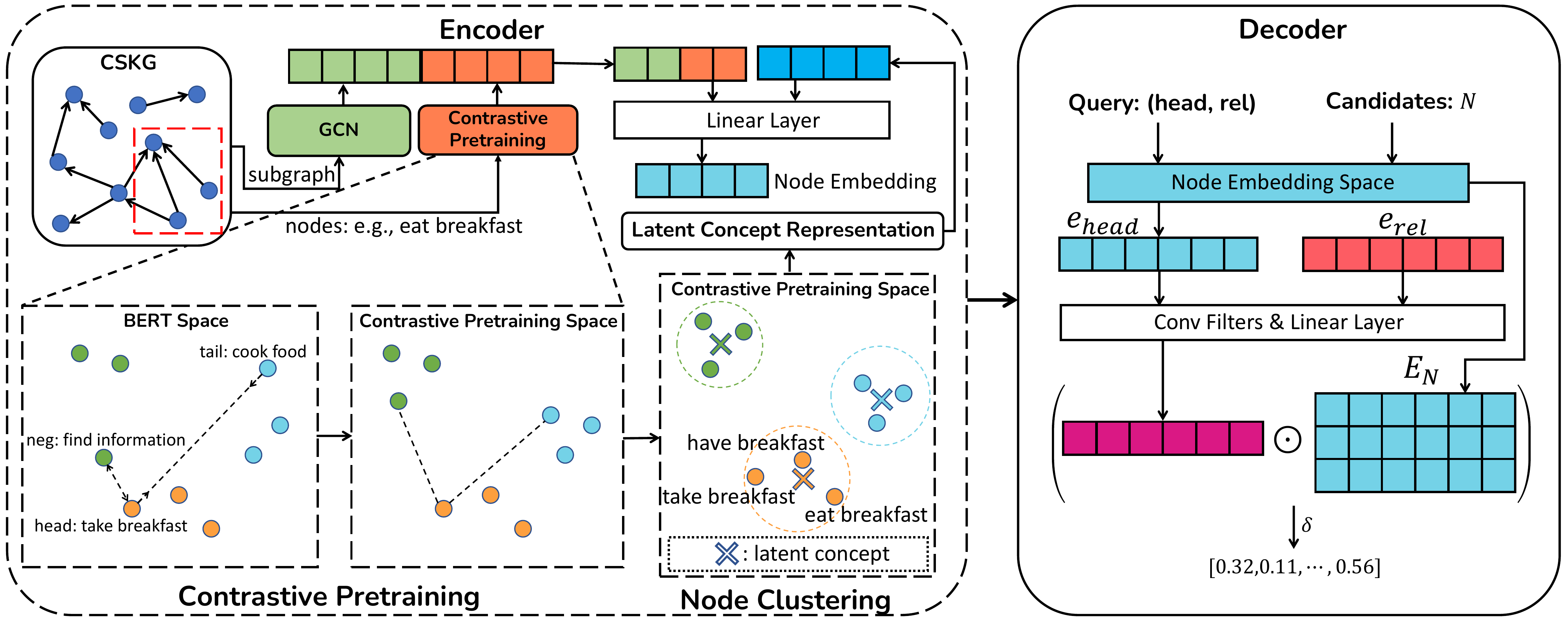}
    \caption{Overall architecture of CPNC. 
    The Encoder is to get node embedding.
    Harnessing the power of GCN and Contractive Pretraining, the Encoder transforms CSKG into rich semantic and structural representations. In detail, the lower left section utilizes the mechanism of Contrastive Pretraining. Then, we describe the node clustering process in the lower right, driven by semantic representations from Contractive Pretraining. Finally, the fusion of semantic, structure, and latent concepts in the upper right yields node embeddings. The Decoder part is to rank the nodes in candidates set by Node Embedding.
    }
    \label{fig:Model}
\end{figure*}

\paragraph{Traditional Knowledge Graph Completion}
Many knowledge graph completion methods have been proposed, which can be classified into three types: embedding-based completion methods, path-finding-based completion methods, and logical rule-based completion methods.
The embedding-based method learned the relation and node embedding by shortening the distance between the head-relation pair representation and the tail node representation, which had good scalability.
Moreover, convolution has been proven to be an effective operation for acquiring the head-relation representation~\cite{DBLP:conf/aaai/DettmersMS018,DBLP:conf/aaai/ShangTHBHZ19} in the embedding-based method.
The path-based methods used the random walk inference based algorithm to find the related path.
They achieved the prediction of missing tuples by comparing the related path with the relation to be predicted~\cite{DBLP:journals/ml/LaoC10,DBLP:conf/icdm/KhotNKS11}.
The rule-based methods utilized the induction rules to simplify the path-finding process in the knowledge graphs~\cite{DBLP:conf/iclr/RenHL20,DBLP:conf/iclr/YangS20}.
We refer the reader to~\citep{DBLP:journals/tnn/JiPCMY22} for more details about knowledge graph completion.

\paragraph{Commonsense Knowledge Graph Completion}
The existing methods assume specific relations, dense edges, and sufficient training samples in the knowledge graph.
However, the sparsity of edges and the abstract nature of relations pose challenges for directly applying these methods to the CSKG.

Early methods in this field employed a strategy where, for a given head and tail, all relations were used to generate a large number of tuples.
BiLSTMs and linear transformations were then utilized to score these tuples, allowing for the prediction of missing tuples in the CSKG~\cite{DBLP:conf/conll/SaitoNAT18,DBLP:conf/acl/LiTTG16,DBLP:journals/corr/abs-1804-09259}.
Building upon this foundation, ~\citet{DBLP:journals/corr/abs-2210-07621} took a step further. They normalized the tail nodes and used RoBERTa to score these tuples. This approach resulted in the curation of Dense-ATOMIC, a CSKG with increased coverage and a richer set of multi-hop paths.

these methods did not take into account structural information and may suffer from computational inefficiency during inference.
Another line of studies attempted to use the translated-based approach to reduce computational consumption \cite{DBLP:conf/aaai/MalaviyaBBC20,DBLP:conf/cikm/JuYL22,DBLP:conf/ijcnn/WangWHYLK21}.
Furthermore, they incorporated GCN for extracting graph representations and fine-tuned BERT in the CSKG to obtain semantic representations.
This integration capitalizes on the complementary nature of these two representations.
In addition, these methods have applied convolutional layers to enhance the performance of CSKG completion. This further improves the effectiveness of the completion task.
However, it was observed that simply fine-tuning BERT~\cite{DBLP:conf/naacl/DevlinCLT19} with pretraining tasks is hard to effectively capture the sentence-level semantic connection between two nodes. In response to this limitation,~\citet{DBLP:journals/corr/abs-2210-07570} proposed MICO, which learns distinct node representations for head nodes under varying relations, proving evidence that high-quality node representation can significantly aid CSKG completion.
Moreover, different nodes in the CSKG may express the same concept. 
The node redundancy will bring difficulties to the commonsense inference on the CSKG~\cite{DBLP:conf/acl/JungPCLKK022}, while the previous CSKG completion methods ignored.

Compared to these methods, we address the issue of node redundancy by employing a clustering algorithm. Our work stands out by learning the semantic representations of nodes through sentence-level semantic connections and integrating latent concept information into node representations.

\section{Task Definition}
Given a CSKG $G=(N,V,R)$ where $N$ is the set of nodes, $V$ is the set of edges and $R$ is the set of relations. We regard each tuple $v_i= (h,rel,t)$ in the CSKG as a sample, which is composed of head node $h$, tail node $t$ and relation $rel$, where $h,t \in N$, $v_i \in V$, and $rel \in R$.
Given a query $(h,rel)$ formed by a head node $h$ and a relation $rel$, the target of the CSKG completion is to maximize the score of the tail node $t$.
Following the previous work \citep{DBLP:conf/aaai/MalaviyaBBC20}, for an edge $(h,rel,t)$ existing in the CSKG, we also add an inverse edge $(t,rel^{-1},h)$ to the graph, where $rel^{-1}$ is the inverse relation of $rel$.

\section{Approach}
In this paper, we propose a new framework, CPNC, for CSKG completion using an Encoder-Decoder architecture (see Figure~\ref{fig:Model}). The Encoder incorporates semantic, graph structure, and latent concept representation obtained from Contrastive Pretraining, Graph Convolutional Network (GCN), and Node Clustering, respectively, to acquire node representations. Given query $(h, rel)$, the Decoder ranks the nodes in the candidate set $N$ and finds tail nodes for the query by the rank.

\subsection{Contrastive Pretraining}
The current mainstream approach for completing CSKG utilizes translation-based methods and relies on GCN to represent the graph structure of nodes.
Although GCNs are effective in modeling graph structure, they have limitations in capturing graph structure information in CSKG due to sparse edges.
To address this issue, these methods incorporate semantic information by fine-tuning BERT on CSKG using Masked Language Model (MLM) and Next Sentence Prediction (NSP) tasks.

However, those methods are imperfect in modeling CSKGs composed of nodes presented as short text.
MLM, which is a token-level task, focuses on modeling the connection between masked words and other words, making it unsuitable for achieving sentence-level node representation.
Similarly, NSP, while capable of modeling the semantic connection between two nodes at the sentence level, suffers from a mismatch between the input of the pretraining phase and the CSKG completion phase. During pretraining, NSP requires a pair of head and tail nodes for prediction, but during CSKG completion, only a single node is used for sentence representation.
Consequently, NSP is not well-suited for the task of learning node representations, and similar observations have been made in the field of sentence representation learning~\cite{DBLP:conf/emnlp/LiZHWYL20}.

To obtain better node representations for the CSKG completion, we introduce Contrastive Pretraining (CP), a new method that sufficiently leverages semantic information at the sentence level.
CP focuses on CSKG completion and differs from MICO~\cite{DBLP:journals/corr/abs-2210-07570} by not incorporating relation categories.
Instead, it fine-tunes BERT's node representation using contrastive learning and capture sentence-level connections between nodes.

\subsubsection{Building Contrastive Learning Samples}
Assuming node pairs linked in the CSKG are semantically related, we consider node pairs without a link in the CSKG as semantically unrelated.

\begin{figure}[!htp]
    \centering
    \includegraphics[width=0.9\columnwidth]{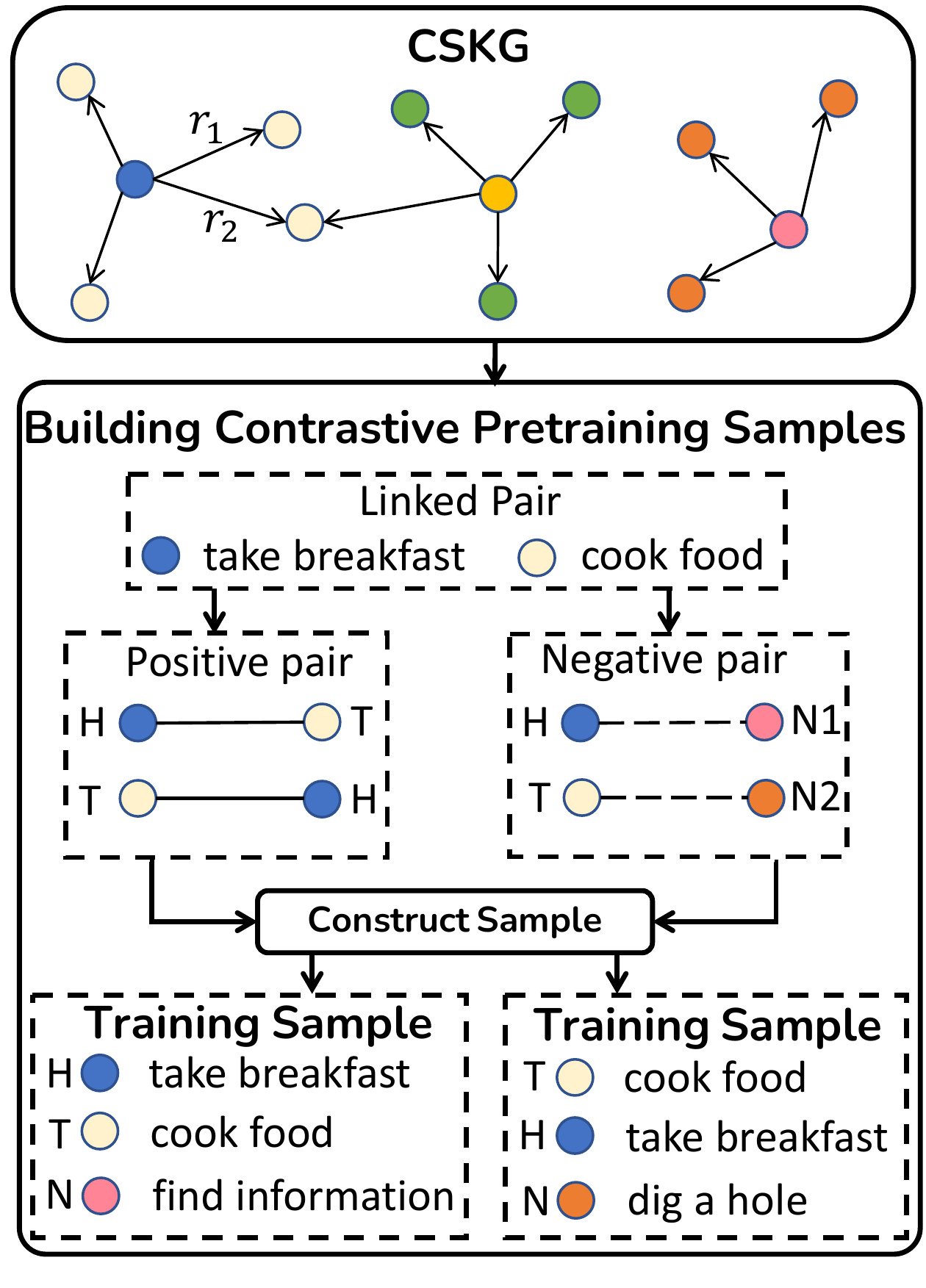}
    \caption{Main process of building Constractive Pretraining samples.}
    \label{fig:Contrastive Sample}
\end{figure}

For each arbitrary edge $(h, rel, t)$ in CSKG, we randomly sample a node $\bar{t}$ having no linking with the head node $h$ as its hard negative tail and construct a contrastive learning training sample $c=({h,t,\bar{t}})$.
As shown in Figure~\ref{fig:Contrastive Sample}, ``take breakfast'' and ``cook food'' are the head and tail nodes in a edge of CSKG.
We randomly select a node ``\emph{find information}'' having no linking with ``\emph{take breakfast}'' to construct a training sample \emph{(``take breakfast'', ``cook food'', ``find information'')}.

The previous methods primarily focused on the head nodes while constructing negative samples, neglecting the importance of capturing unrelated semantics among the tail node set.
To tackle this problem, we additionally constructed a contrastive learning sample $c_{reverse}=({t,h,\bar{h}})$ for the tail node $t$ in the edge $(h, rel, t)$.

\subsubsection{Multiple Negatives Ranking Loss}
We employ BERT for node embedding by inputting all nodes in a batch separately and applying Mean Pooling on the final layer's output.
Then, we use $x(n)$ to denote the representation of a specific node $n$ obtained through BERT.

In order to ensure that semantically related nodes are positioned closely in the embedding space, we utilize contrastive learning and employ the Multiple Negatives Ranking Loss for contrastive learning.  

For a given sample $c_i = (h_i, t_i, \bar{t_i})$ in a batch, we employ the Multiple Negatives Ranking Loss. We consider $h_i$ paired with $t_i$ as the positive sample while treating $h_i$ paired with other tail nodes $t_j$ in the batch as negative samples. Additionally, we consider $h_i$ paired with all hard negative tails in the batch as negative samples as well. The objective is to minimize the semantic distance between the nodes in positive samples within the batch. The formulation of the multiple negatives ranking loss is as follows:

\begin{equation}
 L = -\sum_{i=1}^{M}\log\frac{e^{D(h_i,t_i))}}{\sum_{j=1}^{M}e^{D(h_i,t_j)}+e^{D(h_i,\bar{t}_j)}},
\end{equation}
where $D(h,t)=(x(h) \cdot x(t)^{T})/(\|x( h)\|\cdot\|x(t)\|)$ represents the cosine similarity of two nodes, $M$ is the number of samples in the a batch.

\subsection{Encoder}
The Encoder aims to generate node representations for CSKG completion,as illustrated in the left part of Figure~\ref{fig:Model}.

The node representation comprises two components,

1) the semantic representation matrix of nodes $E_{sem}$ obtained through Contrastive Pretraining:
\begin{equation}
    E_{sem} = \text{Contrastive-Pretraining}(N),
\end{equation}

2) the graph structure representation matrix of nodes acquired using a Graph Convolutional Network (GCN),
\begin{equation}
    E_{graph} = \text{GCN}(G).
\end{equation}

Specifically, $e_{sem}$ and $e_{graph}$ represent the semantic and graph representations of a particular node, respectively, which can be derived from $E_{sem}$ and $E_{graph}$.

\subsection{Node Clustering}
On the basis of the semantic node representation $e_{sem}$, we use the K-means algorithm to cluster all nodes $N$ in the graph as follows:
\begin{equation}
    \{\mathcal{C}_1,\mathcal{C}_2,...,\mathcal{C}_K\}=\text{K-means}(N),
\end{equation}
where $\mathcal{C}_i$ is the $i$-th cluster, and $K$ represents the number of clusters.
Nodes in $\mathcal{C}_i$ are considered to have the same latent concept.
As shown in the Node Clustering part of Figure~\ref{fig:Model}, ``\emph{eat breakfast
}'', ``\emph{have breakfast
}'' and ``\emph{take breakfast
}'' are in the same cluster and have the same latent concept.

We define the representation of the latent concept as the mean of semantic representation of all nodes in a cluster, denoted as $\text{mean}(\mathcal{C}_k)$, where $\mathcal{C}_k$ contains a set of nodes in the same cluster.

To acquire the final node representation, we combine the node semantic representation, $e_{sem}$, the graph representation, $e_{graph}$, and the latent concept representation, $\text{mean}(\mathcal{C}_k)$. This fusion is performed as follows:
\begin{equation}\label{eq:e_lc}
    e=[e_{sem};e_{graph};\text{mean}(\mathcal{C}_k)] \cdot W_{embedding},
\end{equation}
where $W_{embedding}$ is a weight matrix.
The resulting node embeddings are considered as the \textbf{node embedding space}.

Intuitively, Node Clustering enhances CSKG completion by leveraging information from nodes within the same latent concept. 

We also use a progressive masking process, following~\citet{DBLP:conf/aaai/MalaviyaBBC20}, to integrate latent concept information. Initially, the latent concept representation is fully masked, and throughout the first 100 training epochs, it is gradually unmasked, improving the overall performance.

\subsection{Decoder}
The goal of Decoder is to rank the nodes in the candidate set, which is considered as $N$ in our work.

In order to obtain the query representation, we first use a convolutional layer to fuse the relation and head node representation:
\begin{equation}
    q = \text{Conv}(e_h,e_{rel}),
\end{equation}
where $e_{h}$ is the representation of the head nodes in node embedding space obtained from Eq.~\ref{eq:e_lc}, $e_{rel}$ is the representation of relation $rel$ and $\text{Conv}$ is a convolution operation.

Then, we predict a 0-1 distribution vector, based on the representation of the query $q$, indicating the likelihood of each node in the candidate set $N$ becoming the tail node:
\begin{equation}
    p_t=\delta (q W_{conv} E_{N}),
\end{equation}
where $W_{conv}$ is a weight matrix, $E_{N}$ is the representation matrix of candidate nodes in node embedding space obtained from Eq.~\ref{eq:e_lc}, and $\delta$ is a sigmoid function.

\section{Experiments}

\subsection{Experimental Setup}
\subsubsection{Datasets}
We evaluate our CPNC framework on two CSKG completion benchmarks, i.e., CN-100K~\citep{DBLP:conf/aaai/SpeerCH17} and ATOMIC~\citep{DBLP:conf/aaai/SapBABLRRSC19}.

\textbf{CN-100K} is a dataset that encompasses general commonsense knowledge. 
This version contains 36 relation types and the Open Mind Common Sense (OMCS) entries from ConceptNet~\cite{DBLP:series/tanlp/SpeerH13}. The average length of the nodes in CN-100K is 2.85 words.
Following~\citet{DBLP:conf/aaai/MalaviyaBBC20}'s split, the training set contains 10,000 tuples, and the validation set and test set both contain 1200 tuples.

\textbf{ATOMIC} is an atlas of everyday commonsense reasoning and primarily focuses on event-level commonsense knowledge in the form of if-then relations.
It comprises 9 relation types, with an average of 4.40 words per node. 
We split ATOMIC following~\citet{DBLP:conf/aaai/MalaviyaBBC20}'s work, where the training set consists of 610,536 tuples, while the validation and test sets contain 87,700 tuples and 87,701 tuples, respectively.

\begin{table*}
\begin{center}
\footnotesize
 \resizebox{2.1\columnwidth}{!}{
    \begin{tabular}{ll|llll|llll} 
    \hline
    \multicolumn{2}{c|}{\textbf{Methods}} & \multicolumn{4}{c|}{\textbf{CN-100K}} & \multicolumn{4}{c}{\textbf{ATOMIC}} \\ 
    Type & Model & {\sc MRR} & {\sc Hits@1} & {\sc @3} & {\sc @10} & {\sc MRR} & {\sc Hits@1} & {\sc @3} & {\sc @10} \\
    \hline
    \multirow{4}{*}{KG-adapted } &
     DistMult & 
     \ \  8.97 & \ \ 4.51 & \ \ 9.76 & 17.44 &
     12.39 & \ \ 9.24 & 15.18 & 18.30 \\
     ~ & ComplEx & 
     11.40 & \ \ 7.42 & 12.45 & 19.01 &
     14.24 & \textbf{13.27} & 14.13 & 15.96 \\
     ~ & ConvE & 
     20.88 & 13.97 & 22.91 & 34.02 &	
     10.07 & \ \ 8.24 & 10.29 & 13.37 \\
     ~ & ConvTransE  & 
     18.68 & \ \ 7.87 & 23.87 & 38.95 &	
     12.94 & 12.92 & 12.95 & 12.98 \\
     \hline
     \multirow{2}*{Generation-based} &
     COMeT-Normalized  &
     \ \ 6.07 &  \ \ 0.08 &  \ \ 2.92 &  21.17 &
     \ \ 3.36 &  \ \ 0.00 & \ \ 2.15 & 15.75 \\
     ~ & COMeT-Total  &
     \ \ 6.21 & \ \ 0.00 & \ \ 0.00 & 24.00 &	
     \ \ 4.91 & \ \ 0.00 & \ \ 2.40 & 21.60  \\
     \hline
     \multirow{3}*{CSKG-dedicated} &
     RGAT & 
     43.97 & 30.75 & 51.54 & 69.34 &
     \quad - & \quad - & \quad - & \quad - \\
     ~ & SGBC & 
     49.12* & 37.71* & 56.67* & 71.29* &
    10.25* &  \ \ 8.72* & 10.54* & 13.26* \\
     ~ & InductivE & 
     56.92* & 45.54* & 63.38* & 78.63* &
     13.19* & 10.26* & 13.61* & 18.83* \\ 
     \hline
     \multirow{2}*{Ours}
     & CPNC-S & 
     54.52 & 45.33 & 61.46 & 75.92 &
     13.14 & 10.11 & 13.75 & 18.80 \\
     ~ & CPNC-I & 
     \textbf{59.00} & \textbf{48.29} & \textbf{65.04} & \textbf{79.13} &
     \textbf{14.38} & 10.53 & \textbf{15.22} & \textbf{21.79} \\
     \hline
\end{tabular}
 }
\end{center}
\caption{ CSKG completion results on CN-100K and ATOMIC. The result of HITS@1 was not reported in \citet{DBLP:conf/ijcnn/WangWHYLK21}. 
To fairly compare our method with the previous method in the same setting, we rerun the code of InductivE and SGBC on ATOMIC and CN-100K and mark the results of rerun experiments with *. 
Those results are a little lower than that in the original paper.
For KG-adapted and Generation methods, we reuse the results reported in~\citet{DBLP:conf/aaai/MalaviyaBBC20}.
}
\label{tab:Main Result}
\end{table*}

\subsubsection{Evaluation Metric}
We evaluate the performance of our method using MRR and HITS, following previous CSKG completion methods~\cite{DBLP:conf/ijcnn/WangWHYLK21,DBLP:conf/aaai/MalaviyaBBC20,DBLP:conf/cikm/JuYL22}.
The results are reported by averaging over both forward tuples $(h,rel,t)$ and inverse tuples $(t,rel^{-1},h)$.
Moreover, because the nodes in CSKG are represented in free-form text, it is possible for nodes other than the golden tail nodes to be considered reasonable tail nodes.
To address this, we conduct a human evaluation to assess the predictions made by our models.

\subsubsection{Implementation Details}
To perform Contrastive Pretraining, we use the contrastive learning framework provided in
\url{https://github.com/UKPLab/sentence-transformers}.
Our approach employs BERT-large as the base model, which contains 340M parameters.
During training, we utilized a batch size of 128 and conducted training for 3 epochs on the CN-100K and ATOMIC datasets. 
We choose the Adam optimizer for optimization, setting the learning rate to 1e-4 for BERT-large and 5e-5 for the MLP.
For the remaining hyperparameters, we used the default values provided by the framework.

For the CPNC-I and CPNC-S model, we use experimental settings proposed by \citet{DBLP:conf/aaai/MalaviyaBBC20} and \citet{DBLP:conf/ijcnn/WangWHYLK21}, respectively.
Both models are trained for a minimum of 200 epochs using BERT-large, which consists of 340M parameters, to encode semantic representations.
During training, we evaluate the MRR on the development set every 10 epochs for CN-100K and ATOMIC. Training continues until no further improvement in MRR is observed.
We select the model checkpoint that achieves the highest MRR on the development set for testing.

\subsection{Compared Systems}
we compare our approach with nine baseline systems across three categories.

\subsubsection{KG-adapted Methods}
We adapt classic knowledge graph completion methods for CSKG completion.
\textbf{DistMult}~\citep{DBLP:journals/corr/YangYHGD14a} employed a bi-linear product to calculate score of a tuple;
\textbf{ComplE}~\citep{DBLP:conf/icml/TrouillonWRGB16} utilized imaginary number representation to effectively handle a large number of relations in the knowledge graph;
\textbf{ConvE}~\citep{DBLP:conf/aaai/DettmersMS018} fused the representation of the source node and the relation through a 2D convolution layer to obtain the representation of $\rm query$ $\rm(h,rel)$;
\textbf{ConvTransE} extended ConvE by incorporating the translational properties of TransE.

\subsubsection{Generation-based Methods}
\textbf{COMeT}~\cite{DBLP:conf/acl/BosselutRSMCC19}  is a Transformer-based knowledge generation model.
Following~\citet{DBLP:conf/aaai/MalaviyaBBC20}, we adapt COMeT for the CSKG completion and only evaluate in the forward direction. Additionally, we also use their modifications to the ranking method used in COMeT. Specifically, \textbf{COMET-Normalized} and \textbf{COMET-Total} use the normalized negative log-likelihood scores and total log-likelihood scores, respectively, to rank nodes in the candidate set.

\subsubsection{CSKG-dedicated Methods}
SIM+GCN+BERT+ConvTransE (\textbf{SGBC}) densified the CSKG by connecting the synthetic edges between similar semantic nodes, improving the graph structure representation~\cite{DBLP:conf/aaai/MalaviyaBBC20}.
Moreover, they used fine-tuned BERT to encode the semantic information of the nodes.

\textbf{InductivE}~\citep{DBLP:conf/ijcnn/WangWHYLK21} is proposed to enhance the unseen entity representation with neighboring structural information by densifying Graph.
Relational graph attention networks (\textbf{RGAT}) are proposed weighted the importance of neighbor nodes of each node to obtain a better node representation.

\subsubsection{Our CPNC Methods}
By incorporating CPNC with two CSKG-dedicated methods (SGBC and InductivE), we obtain two models, i.e., \textbf{CPNC-S} and \textbf{CPNC-I}.

\subsection{Main Results}
In Table~\ref{tab:Main Result}, we report experimental results of KG-adapted methods, Generation methods, CSKG-dedicated methods and our methods on CN-100K and ATOMIC.

On CN-100K, both KG-adapted methods and Generation methods exhibit unsatisfactory performance, with MRR values below 21\% and HITS@1 values below 14\%. These results demonstrate that directly applying KG-adapted and Generation methods to CSKG completion is ineffective.
In contrast, CSKG-dedicated methods achieve significantly better results on CN-100K, with MRR values above 43\% and HITS@1 values above 30\%. These metrics are twice as high as those obtained by KG-adapted and Generation methods, highlighting the advantage of CSKG-dedicated methods.
Comparing our proposed CPNC-S and CPNC-I models with mainstream CSKG-dedicated methods (SGBC and InductivE), we observe performance improvements of 5.40\% and 2.08\% on MRR, respectively. Notably, our CPNC-I model sets a new state-of-the-art result on CN-100K.

On ATOMIC, we draw a similar conclusion. The KG-adapted methods exhibit poor performance, while CSKG-dedicated methods achieve better results.
In addition, compared with the SGBC and InductivE models, our CPNC-S and CPNC-I models outperform the SGBC and InductivE models by 2.89\% and 1.19\% on MRR, respectively. 
And the CPNC-I model achieves the state-of-the-art result on ATOMIC.

We conduct Paired t-Test and the result proves that the improvement of CPNC is significant.

\subsection{Ablation Study}
\begin{table}[!htp]
    \centering
    \footnotesize
    \small
    \begin{tabular}{ccccc}
        \hline
         & & MRR & HITS@10 \\ \hline
        \multirow{8}{*}{CN-100K} & \textbf{CPNC-S} & \textbf{54.52} & \textbf{75.92}  \\ 
        ~ & -w/o CP & 51.90 & 76.08  \\ 
        ~ & -w/o NC & 49.74 & 71.38  \\
        ~ & SGBC & 49.12 & 71.29  \\
        \cline{2-4}
        ~ & \textbf{CPNC-I} & \textbf{59.00} & \textbf{79.13}  \\
        ~ & -w/o CP & 57.16 & 74.90  \\ 
        ~ & -w/o NC & 58.30 & 78.75  \\
        ~ & InductivE & 56.92 & 78.63  \\
        \hline
        \multirow{8}{*}{ATOMIC} & \textbf{CPNC-S} & \textbf{13.14} & \textbf{18.80}  \\ 
        ~ & -w/o CP & 12.77 & 17.15  \\ 
        ~ & -w/o NC & 12.46 & 17.72  \\
        ~ & SGBC & 10.25 & 13.26  \\
        \cline{2-4}
        ~ & \textbf{CPNC-I} & \textbf{14.38} & \textbf{21.79}  \\
        ~ & -w/o CP & 13.22 & 19.16  \\ 
        ~ & -w/o NC & 14.21 & 21.38  \\
        ~ & InductivE & 13.19 & 18.83  \\
        \hline
    \end{tabular}
    \caption{MRR and HITS@10 of removing CP and NC from CPNC on CN-100K and ATOMIC.}
    \label{tab:Ablation Result}
\end{table}

To demonstrate the effectiveness of Contrastive Pretraining (CP) and Node Clustering (NC) on CSKG completion, we perform the ablation study where we removed CP and NC. The results are presented in Table~\ref{tab:Ablation Result}.

When NC was removed, we can observe that the performance of both CPNC-S and CPNC-I decreases on CN-100K and ATOMIC.
This indicates that NC's latent concept representation effectively assists CSKG completion.
It is worth noting that CP can still bring improvement compared with the previous method, showing that the semantic representation provided by CP is crucial for CSKG completion. 

CPNC, which combines NC and CP, consistently yields the highest result.
By replacing CP with finetuned BERT~\citep{DBLP:conf/aaai/MalaviyaBBC20}, the performance of both CPNC-S and CPNC-I drops. However, they still achieve comparable results with the previous CSKG-dedicated methods.
This demonstrates that the CP requires a high-quality semantic representation of nodes to acquire good latent concept representation.

\subsection{Human Evaluation}
For a given query $ (h, rel)$, additional nodes besides the golden tail node could also become reasonable tail nodes.
For example, given a query ``\emph{(do housework, Causes) }'', besides golden tail node ``\emph{ clean the house}'', ``\emph{ house get clean}'' and ``\emph{ clean the room}'' are also reasonable tail nodes. 
However, automated metrics fail to cover these nodes.
To address this issue, we randomly select 200 queries from the test set of the CN-100K and ATOMIC.
For each query, we use our methods and current mainstream methods to rank the nodes in the candidate set, We manually calculate the number of reasonable tail nodes in the top 10 candidates.
The results are shown in Table~\ref{tab:Human evaluation}.

\begin{table}[!htp]
    \centering
    \small
    \footnotesize
    \begin{tabular}{lcc}
        \hline
          & ATOMIC & CN-100K \\ \hline
        CPNC-I & \textbf{70.22} & \textbf{83.20}  \\ 
        InductivE & 64.45 & 76.40  \\ 
        CPNC-S & 60.23 & 69.40  \\
        SGBC & 62.11 & 65.40  \\ 
        \hline
    \end{tabular}
    \caption{Average percentage of reasonable tuples by human evaluation results.}
    \label{tab:Human evaluation}
\end{table}

Observing human evaluation results, the CPNC model has an excellent performance in CSKG completion.
Besides, our method is significantly higher than the current mainstream methods under the human evaluation results, which further validates the effectiveness of our approach.

\subsection{Model Adaptability}
The CPNC can be effectively applied to various methods.
In our experiment, we incorporate CPNC into SGBC and InductivE, resulting in significant improvements.

The SGBC method initializes the input of GCN randomly and combines graph structural representation acquired by GCN and semantic representation obtained by BERT as the node representation;
InductivE initializes the input of GCN with BERT and uses the output of GCN as the node representation.
Despite the differences in their structures, both methods benefit from the application of CP and NC.

On CN-100K, CPNC-S and CPNC-I bring improvements of 5.40\% and 2.08\% in MRR, respectively;
On ATOMIC, CPNC-S and CPNC-I yield MRR improvements of 2.89\% and 1.19\%, respectively.
This result demonstrates the adaptability of our method.

\subsection{Discussion on the Effect of Relieving Sparsity}
To evaluate the effectiveness of CPNC in mitigating the impact of edge sparsity in the CSKG, we conducted a comparative study between CPNC-S and SGBC on CN-100K with varying sparsity levels. The removal of edges in the CSKG results in increased sparsity, with a higher number of deleted edges corresponding to a higher sparsity degree.

\begin{figure}[!htp]
    \centering
    \includegraphics[width=0.9\columnwidth]{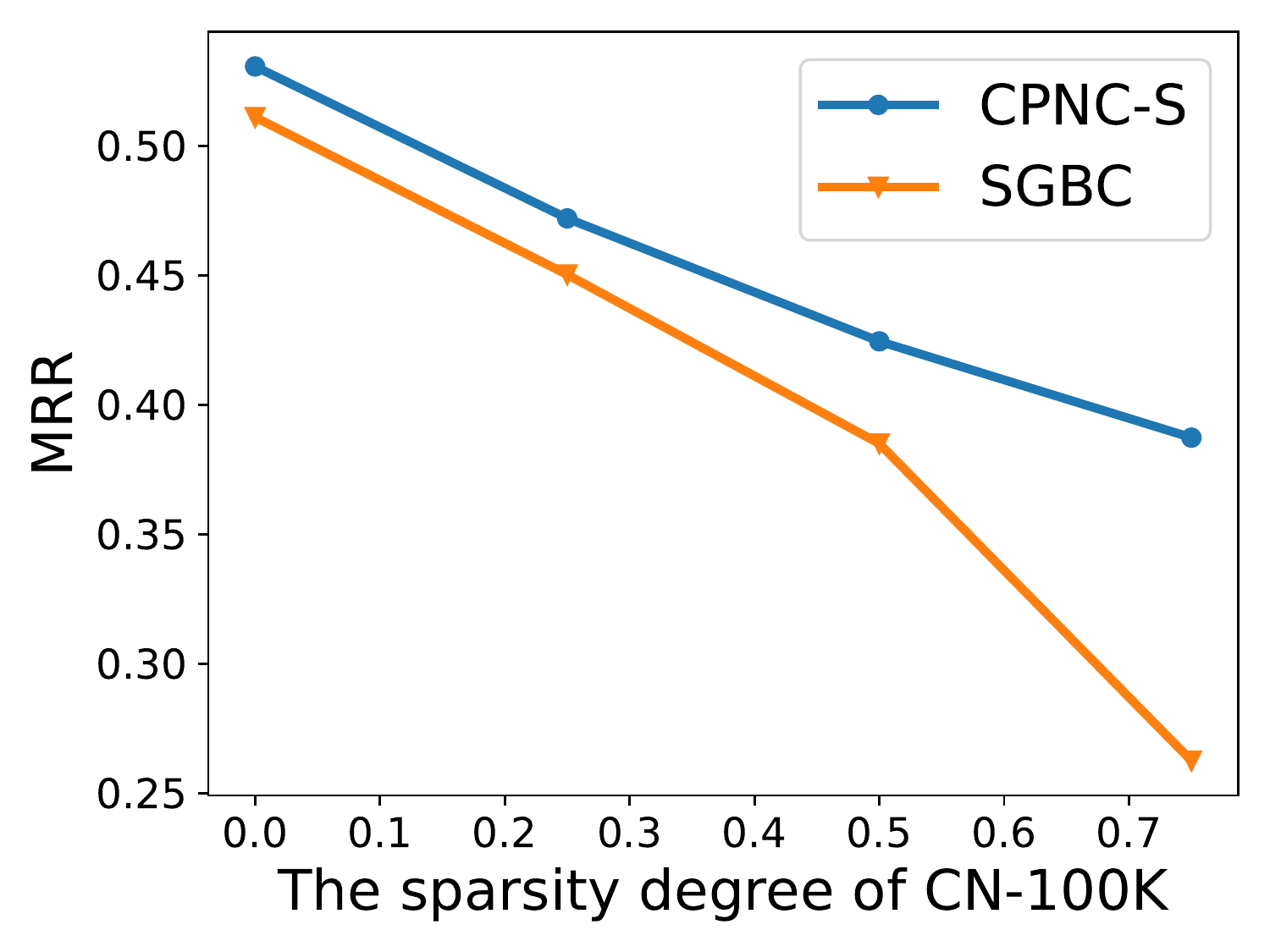}
    \caption{CSKG completion results of SGBC and CPNC-S on CN-100K with various degrees of sparsity.}
    \label{fig:sparity}
\end{figure}

As shown in Figure~\ref{fig:sparity}, when the sparsity degree increases, CSKG completion performance deteriorates.
CPNC-S consistently outperforms SGBC in terms of MRR on CN-100K across different sparsity levels, indicating CPNC's ability to mitigate edge sparsity in the CSKG. CPNC-S achieves a 12.48\% and 3.95\% increase in MRR compared to SGBC at sparsity extents of 25\% and 50\% respectively. These results demonstrate that CPNC is more effective in alleviating edge sparsity in the CSKG, with greater improvements observed for sparser graphs.

\subsection{Discussion on the Number of Latent Concepts}

\begin{figure}[!htp]
    \centering
    \includegraphics[width=0.9\columnwidth]{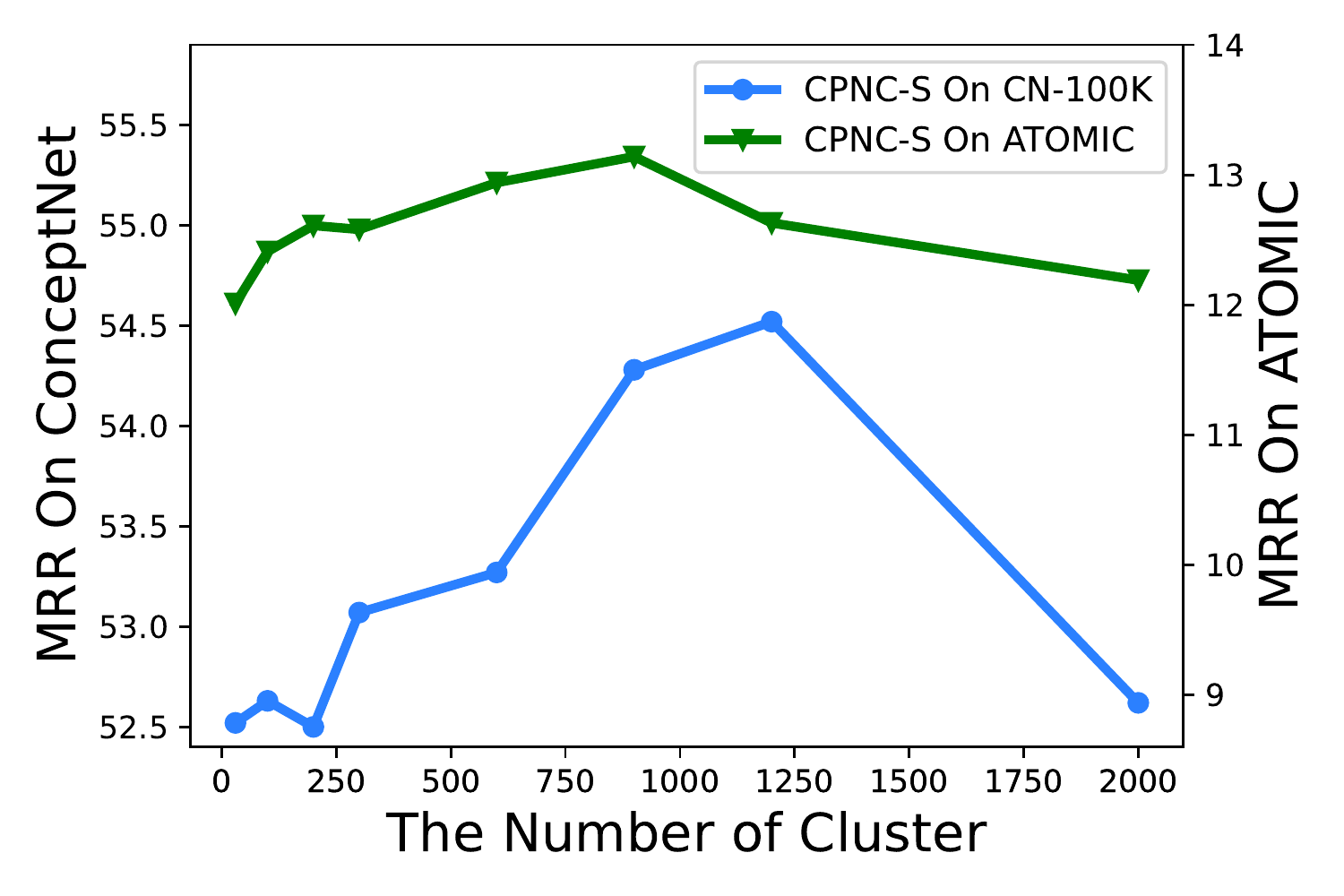}
    \caption{CSKG completion performance with different cluster numbers on CN-100K and ATOMIC.}
    \label{fig:cluster number}
\end{figure}

Different numbers of clusters can lead to varying levels of granularity in latent concepts, which in turn can impact the performance of CSKG completion.
In order to investigate this, we conducted experiments with the different numbers of clusters on CN-100K and ATOMIC datasets and present the results in Figure~\ref{fig:cluster number}.

On CN-100K, we observed an overall increasing trend in completion results as the number of clusters increased. However, the rate of improvement tends to diminish after reaching 900 clusters. The maximum MRR achieved when using the number of clusters is 1200.

On ATOMIC, the MRR initially increases with an increasing number of clusters but starts to decline once the number of clusters reaches 900.

Based on our main results, we selected 1200 clusters for CN-100K and 900 clusters for ATOMIC, allowing us to effectively capture latent concepts through cluster nodes.

\subsection{Discussing on Contrastive Learning Loss}
In order to obtain a better semantic representation of nodes, we experiment with five kinds of contrastive learning methods and observe their performance in CSKG completion.
The result is shown in Table~\ref{tab:Contrastive Loss Choose}.

\begin{table}[!htp]
    \centering
    \scriptsize
    \small
    \resizebox{0.9\columnwidth}{!}{
        \begin{tabular}{cc}
            \hline
            Methods & MRR \\ 
            \hline
            Contrastive Loss & 40.03  \\ 
            MTriplet Loss & 44.36  \\ 
            MICO & 45.93  \\
            Batch Semi Hard Triplet Loss & 27.92  \\ 
            \textbf{Multiple Negatives Ranking Loss} & \textbf{49.74}  \\ 
            \hline
        \end{tabular}
    }
    \caption{CSKG completion performance of CPNC-S using different Contrastive Learning Loss on CN-100K.}
    \label{tab:Contrastive Loss Choose}
\end{table}
In order to compare those contrastive learning methods, we use the node embedding obtained by those methods to complete CSKG without node clustering.
Each contrastive learning method yields distinct results, and our main result selects Multiple Negatives Ranking Loss due to its highest MRR.

\section{Conclusion}
In this work, we propose a new CSKG completion framework CPNC to address issues arising from node redundancy and edge sparsity.
CPNC obtains better semantic node information through Contrastive Pretraining, which alleviates the problems caused by edge sparsity.
CPNC also utilizes the latent concept representation acquired through Node Clustering to alleviate the problem caused by node redundancy.
On CN-100K and ATOMIC, experimental results and extensive analysis demonstrate the effectiveness of Contrastive Pretraining and Node Clustering.

\section*{Limitations}
Due to the limitation of time and resources, in this work, we select a relatively small number of clusters during the clustering process, which results in coarse-grained clustering.
Fine-grained clustering can provide a better latent concept but will also lead to increased computational resources and time consumption.
We will attempt to trade-off between the cost and the granularity of clustering in future work to further explore the impact of the latent concept on CSKG completion.
Besides, our Node Clustering module is not integrated in an end-to-end manner in our work; We will consider using the topic neural network to construct an end-to-end model.

\section*{Ethics Statement}
We would like to thank \citet{DBLP:conf/aaai/MalaviyaBBC20} and \citet{DBLP:conf/ijcnn/WangWHYLK21} for their code on commonsense knowledge graph completion. 
Their models are licensed under MIT, which allows copying, modifying, merging, publishing, and distributing of the material.

In the stage of human evaluation, we employed three graduate students experienced in natural language processing for human evaluation. We paid the graduate students about \$8 per hour, well above the local average wage, and engaged in constructive discussions if they had concerns about the process.

\section*{Acknowledgements}
This work was supported by the Natural Science Foundation of China (No. 62076133), and the Natural Science Foundation of Jiangsu Province for Distinguished Young Scholars (No. BK20200018).

\bibliography{acl2023}
\bibliographystyle{acl2023}

\end{document}